\documentclass[journal,comsoc]{IEEEtran}
\usepackage{epsfig}

\usepackage[T1]{fontenc}

\ifCLASSINFOpdf
\else
\fi

\usepackage{amsmath}

\begin{document}

\title{Bias-corrected estimator for intrinsic dimension and differential entropy--a visual multiscale approach}

\author{Jugurta~Montalv\~ao,
        ~J\^anio~Canuto,
        ~Luiz~Miranda}

\markboth{ } 
{Shell \MakeLowercase{\textit{et al.}}: Bias-corrected estimator for intrinsic dimension and differential entropy--a visual multiscale approach}

\maketitle

\begin{abstract}
Intrinsic dimension and differential entropy estimators are studied in this paper, including their systematic bias. A pragmatic approach for joint estimation and bias correction of these two fundamental measures is proposed. Shared steps on both estimators are highlighted, along with their useful consequences to data analysis. It is shown that both estimators can be complementary parts of a single approach, and that the simultaneous estimation of differential entropy and intrinsic dimension give meaning to each other, where estimates at different observation scales convey different perspectives of underlying manifolds. Experiments with synthetic and real datasets are presented to illustrate how to extract meaning from visual inspections, and how to compensate for biases.
\end{abstract}


\IEEEpeerreviewmaketitle

\section{Introduction}
\label{Sec:Introduction}

\IEEEPARstart{I}{ntrinsic} dimension (ID) estimation is a useful tool whenever patterns presented in $D$-dimensional spaces are supposed to form structures (manifolds) in $d$-dimensional subspaces, with $d<D$. Examples of such lower dimensional structures are: projections of a rigid objects whose pictures, with $D$ pixels, are taken under $d$ degrees of freedom \cite{Rozza2012, Tenenbaum2000}, or $D$-dimensional representations of vowel sounds, whereas the vocal tract that generates the sound has only $d$ mechanical degrees of freedom \cite{Nilsson2007}. 

In all those applications, if probabilistic models are used to represent the source of observations (i.e., the underlying $d$-dimensional structures) then entropy, differential entropy (DE) and entropy rate \cite{Shannon1948} can reveal relevant attributes of the corresponding structures. In pattern recognition, estimating both ID and DE is tantamount to analysing shape attributes of manifolds, as explained in Section \ref{Sec:ID_DE},  thus suggesting tools for proper design and analysis of classifiers, in special those based on autoencoders. Indeed, while the number of deep neural network applications increases at an astonishing pace, some attempts to explain this success seem to suggest that most answers come from the study of physical restrictions \cite{Lin2017} and consequent formation of data manifolds \cite{Brahma2016,Lecun2015,Yu2018,Montufar2014}.

Although ID does not impose a probabilistic model to be estimated, many published ID estimators are based on probabilistic reasoning \cite{Camastra2016,Campadelli2015,Costa2004Learning,Facco2017,Grassberger1983_1,Levina2005,Little2009}. Indeed, even the well known Grassberger-Procaccia (GP) estimator \cite{Grassberger1983_1}, whose aim is to characterize strange attractors in dissipative (deterministic) dynamical systems, also uses the information-theoretic framework to better explain the kind of {\em dimension} their method is able to estimate (also referred to as {\em information dimension}). 

The formulation proposed in \cite{Grassberger1983_1} includes the use of random variables (RV) as the source model for observations, and explicitly shows a link between intrinsic dimension and differential entropy. Some subsequent works followed this same path, such as \cite{Nilsson2007} and the series of publications by Costa and Hero \cite{Costa2004Learning,Costa2006Determining,Costa2004Geodesic}. However, most published works deal either with DE, under the assumption that ID is known, or with ID estimation, regardless the manifold's volume (thus its DE, as explained in  Section \ref{Sec:ID_DE}). Indeed, in \cite{Nilsson2007} it is stated that
\begin{quotation}
{\em The existence of manifold structures in the data is often overlooked
in entropy estimations, with the result that classical methods,
assuming the wrong intrinsic dimension (manifold dimension)
provide erroneous estimates of the entropy.}
\end{quotation}

On the other hand, in \cite{Ma1981}, the problem of DE estimation in high-dimensional spaces was tackled through a simple but data-efficient approach, referred to as the Coincidence Method (CM), originally applied in Physics. In \cite{Montalvao2014} this method was extended to differential entropy estimation in the pattern recognition context, which clearly shows that the {\em correlation dimension} in \cite{Grassberger1983_1} uses the same empirical coincidence ratio as the entropy estimation method proposed in \cite{Ma1981}.

More specifically, the correlation integral defined in \cite{Grassberger1983_1} is equivalent to the inverse of the number of coincidences defined in \cite{Ma1981}. This equivalence is even more striking in non-redundant reformulations of the correlation integral, as in \cite{Golay2015}. This suggests a link between works from different domains, developed in this paper to yield a visual method where ID and DE are regarded as complementary parameters of the same estimation problem. 

Unfortunately, both methods \cite{Grassberger1983_1,Ma1981} yield biased estimates, a distortion whose source is also shared by them, which is explained by their common theoretical ground. Concerning the bias in the GP method, a theoretical model was first proposed in \cite{Smith1988}, where it was shown that ID bias can be predicted on average if the actual ID is known. In this paper, the theoretical model proposed in \cite{Smith1988} is developed to the point of predicting and compensating for both ID and DE biases, even if the actual ID is unknown.   

This paper is organized as follows: In Section \ref{Sec:ID_DE}, we present a brief recall of ID and DE, and their complementary meanings, whereas in Section \ref{Sec:Approach} the theoretical foundations of the joint estimator proposed in this paper are presented. Finally, in Section \ref{Sec:Proposal}, the method is presented, along with a bias compensation approach. Experiments with both real and artificial data are presented in Section \ref{Sec:Results}. We discuss the main contributions of this work in Section \ref{Sec:Conclusions}. 

\section{Intrinsic dimension and differential entropy in a nutshell}
\label{Sec:ID_DE}

According to \cite{Camastra2002}, the ID of a given set of observations is {\em ``the minimum number of free variables
needed to represent the data without information loss''}, which agrees with the definition in \cite{Bennett1965}, where {\em ``the intrinsic dimensionality of a collection of signals is defined to be equal to the number of free parameters required in a hypothetical signal generator capable of producing a close approximation to each signal in the collection}''. 

DE, on the other hand, is defined as the entropy of a continuous random variable \cite{Cover2012}. Besides, \cite{Hill1973} presents entropy as an effective cardinality in logarithmic scale. Likewise, DE can be regarded as an effective volume (in logarithmic scale) \cite{Cover2012,Montalvao2014}. 

For a brief recall on ID and DE, consider the data sources labeled `Sinusoid' and `Circle', borrowed from \cite{Hein2005}. Although experiments there just consider ID, these datasets can also be used to address DE as well. These sources are defined respectively as
\[
{\bf X}_{Sin}=[\sin(2 \pi U),~\cos(2 \pi U),~0.1\sin(300 \pi U)]
\]
\noindent and
\[
{\bf X}_{Cir}=[\sin(2 \pi U),~\cos(2 \pi U),~0.1 V]
\]
\noindent where $U$ and $V$ are independent random variables uniformly distributed between 0 and 1.

\begin{figure}[htb]
\centering{\includegraphics[width=80mm]{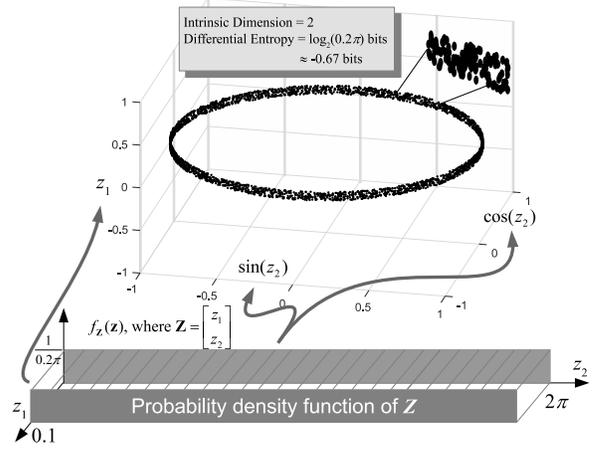}}
\caption{Dots represent instances of ${\bf X}_{Cir}$, which are generated from instances of a 2-D uniform latent random variable ${\bf Z}=[0.1V; ~2 \pi U]$. Thus the intrinsic structure of ${\bf X}_{Cir}$ is planar ($d=2$), in spite of its 3D ($D=3$) representation. The DE of ${\bf X}_{Cir}$ is given by the surface area ($0.2\pi$), thus $h({\bf X}_{Cir})= \log_2(0.2\pi) \approx -0.67$ bits.}
\label{Fig01}
\end{figure}

\begin{figure}[htb]
\centering{\includegraphics[width=80mm]{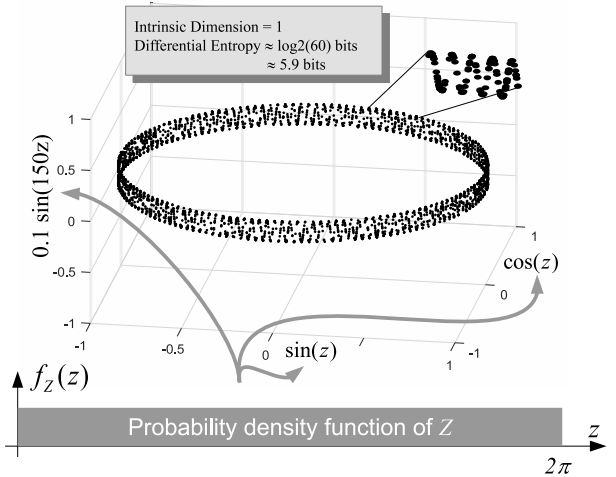}}
\caption{Dots represent instances of ${\bf X}_{Sin}$, which are generated from instances of a single uniform latent random variable $Z=2 \pi U$. Thus the intrinsic structure of ${\bf X}_{Sin}$ is 1D ($d=1$), in spite of its 3D ($D=3$) representation. The DE of ${\bf X}_{Sin}$ is given by the effective structure length (approximately 60), thus $h({\bf X}_{Sin}) \approx 5.9$ bits.}
\label{Fig02}
\end{figure}

Figures \ref{Fig01} and \ref{Fig02} show 3000 instances of ${\bf X}_{Cir}$ and ${\bf X}_{Sin}$ respectively. Their ID are 2 and 1, for the domain of ${\bf X}_{Sin}$ can be cut and straightened to a line segment (1D) of length slightly greater than $60$, whereas the domain of ${\bf X}_{Cir}$ can also be cut and unbent to a rectangle (2D) of area $0.2\pi$.

If the probability density function (pdf) of an RV is known, its R\'enyi $\alpha$-entropy \cite{Cover2012} can be obtained as
\[
h_\alpha ({\bf X}) = \frac{1}{1-\alpha} \log_2 \int_{{\mathcal R}^D} (f_{\bf X}({\bf x}))^\alpha d{\bf x} 
\]
\noindent where $d{\bf x}$ is a differential hypervolume in ${\mathcal R}^D$, only taken  where the pdf $f_{\bf X}({\bf x})$ is not null. Therefore, if the pdf is not null in $d$-dimensional manifolds ($d<D$) the integral must be restricted to it (therefore the local dimensions of the manifolds must be known). This definition encompasses both Shannon DE, for $\alpha \rightarrow 1$, and quadratic (or collision) entropy, for $\alpha = 2$. In both cases, $h_\alpha ({\bf X})$ can also be regarded as a proportion between volumes, suggesting that DE is a measure of {\em effective volume} for non-uniform distributions, as much as entropy is presented as an {\em effective cardinality} for discrete RVs \cite{Hill1973,Montalvao2014}. 

This intuitive perception of DE can be better explained with the notion of effective length, area or volume, as follows: an RV defined as $Z=\lambda U$ ($\lambda \in {\mathcal R}$ and $\lambda >0$) is uniformly distributed along an 1D domain of length $\lambda$, then its DE is given by this length $\lambda$ measured in logarithm scale, $h(Z)=\log(\lambda)$. In general, for non-uniform RV, the effective hypervolume is given by the hypervolume associated to another uniformly distributed RV whose observation removes the same amount of uncertainty about the outcome \cite{Shannon1948}.

Both formal and intuitive points of view reveal a tricky aspect of DE estimation, that the DE is meaningless before the ID is known. Figures \ref{Fig01} and \ref{Fig02} can be used to further illustrate this point, because both RV ${\bf X}_{Cir}$ and ${\bf X}_{Sin}$ are defined in a 3D space, but they have ID equal to 2 and 1, respectively. Therefore, the DE associated to ${\bf X}_{Cir}$  must take a unit square as area reference to yield $h({\bf X}_{Cir})=\log_2(0.2\pi)$ bits, whereas ${\bf X}_{Sin}$ must take a unit line segment as length reference to yield $h({\bf X}_{Sin}) \approx \log_2(60)$ bits. In both cases, an observer unaware of these IDs would fail to estimate the DE, because both datasets are presented as 3D patterns, but their underlying structures have null volume.

\section{Joint analysis approach}
\label{Sec:Approach}

In this Section, we briefly recall two known approaches for ID and DE estimation that, when put side by side, reveal their striking equivalences. These equivalences are then articulated to yield a joint visual analysis for ID and DE.

\subsection{Intrinsic dimension estimation}
\label{Sec:IDestim}
Given a set of $N$ observations, $\{ {\bf x}(1),{\bf x}(2),\ldots, {\bf x}(N)\}$, and a threshold $r$, the ``information dimension'' (also known as correlation dimension) is defined in \cite{Grassberger1983_1}, and can be obtained from the proportionality
\begin{equation} \label{EqID}
C(r) \propto r^d
\end{equation}
\noindent as $r \rightarrow 0$, where the non-redundant \cite{Golay2015,Smith1988} definition of the correlation integral $C(r)$ is 
\begin{equation}\label{Eq:C}
C(r) = \frac{1}{N(N-1)}\sum_{i<j}^N I({||{\bf x}(i) - {\bf x}(j)||\leq r)}
\end{equation}
\noindent where $I$ is an indicator function, i.e. $I(\lambda)=1$ if $\lambda$ is true, and $I(\lambda)=0$ otherwise.

Function  $I(\cdot)$ is a coincidence detection function that allows for the use of any pattern matching measure, or even mean opinion scores, which can be particularly useful for ID estimations in psychometrics or econometrics, for instance. In \cite{Smith1988} the supremum norm is used instead of the original Euclidean norm \cite{Grassberger1983_1}, thus easing theoretical calculations regarding correlation dimension limits.
 
For both definitions, since the volume where coincidence occurs in the manifold scales with $r^d$ (instead of $r^D$), then the number of observation pairs coinciding in this volume should scale at a known rate, if the observation volume is such that the probability density of observations is almost constant inside it.

From Eq. \ref{EqID}, it follows that
\begin{equation} \label{EqID2}
\log C(r) \approx d{\log (r)} - h,
\end{equation}
\noindent where, as $r \rightarrow 0$,  $h$ is the logarithm of the proportionality constant. 

\begin{figure}[htb]
\centering{\includegraphics[width=70mm]{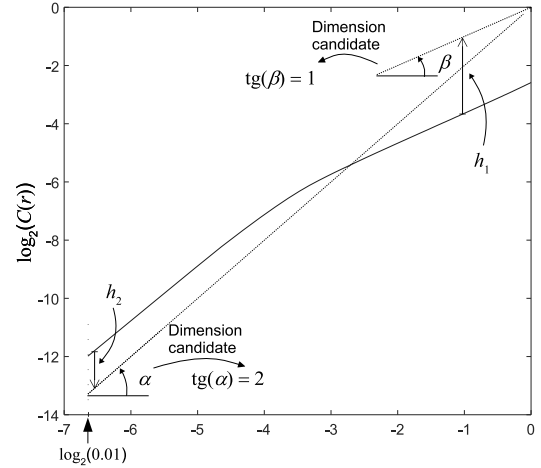}}
\caption{Illustration of the ID estimation method by Grassberger and Procaccia, for 3000 random instances of ${\bf X}_{Cir}$. Depending on the value of $r$, two main linear trends are noticed, thus suggesting two possible ID estimates.}
\label{Fig03}
\end{figure}

To estimate $d$ from Eq. \ref{EqID2}, a common approach is to use the angular coefficient of the line that best fits points $(\log r,~\log C(r))$ in a given range for $r$. Therefore, a single best fit is expected. However, Figure \ref{Fig03} illustrates a case where this expectation is frustrated. This Figure was obtained with $N=3000$ independent observations of ${\bf X}_{Cir}$, and $r$ ranging from 0.01 to 1 (points were interpolated to improve visualization). It may be seen that there are two almost linear intervals with angular coefficients close to either 1 or 2, depending on the range of $r$.

Because GP method is based on results for vanishing values of $r$, one should assume that the estimated ID is 2, corresponding to the lower part of the curved line in Fig. \ref{Fig03}.  Indeed, the detail presented in Fig. \ref{Fig01} clearly shows a 2D local structure. But the estimation for higher values of $r$ is also meaningful, revealing that in a larger scale the 2D structure becomes negligible, whereas an 1D structure emerges. 

That is to say that, on one hand this ambiguity is a drawback of this ID estimator, because bad choices for $r$ may yield inconsistent estimates, whereas good choices remain an open problem \cite{Camastra2016}. On the other hand, this sensitivity to $r$ can be carefully crafted as a tool for multiscale analysis, as discussed in Subsection \ref{Sec:DEestim}.

\subsection{Differential entropy estimation}
\label{Sec:DEestim}
As for differential entropy, our starting point is the estimator proposed by S. Ma in the context of Statistical Mechanics \cite{Ma1981}. This method was motivated by the huge number of reachable physical states in the original problem S. Ma addressed. By replacing states with multivariate random observations, or vectors in an abstract signal space \cite{Bennett1969} we obtain a DE estimator well suited for pattern recognition problems where the amount of observations is small, as compared to the effective size (effective in the sense of \cite{Hill1973}) of the observation domain \cite{Nemenman2011}.   

To estimate the diferential entropy, $h_{\bf X}({\bf x})$, of a random source modelled as ${\bf X}$, we can summarize Ma's method in the following steps:
\begin{itemize}
\item[1.] Arbitrarily set a small hypercube volume $r^d$. It is to be noticed that in the original formulation no intrinsic dimension is considered. Here, however, we consider the possibility of data lying in a manifold of dimension $d \leq D$, which yields an actual hypervolume $r^d \leq r^D$.
\item[2.] Compare all $N_t = N(N-1)/2$ instance pairs ${\bf x}(i)$ and ${\bf x}(j)$, $i < j$, and compute $n_c(r)$ as the number of detected coincidences. A coincidence occurs when $\|{\bf x}(m)-{\bf x}(n)\|_\infty < r/2$.
\item[3.] Compute the ratio between the number of comparisons and the number of coincidences: $Q(r)=\frac{N_t}{n_c(r)}$.
\item[4.] Estimate the {\em effective volume} \cite{Cover2012} of an equivalent uniform pdf as ${\hat V}_{Ma} =  r^d Q(r)$.  
\item[5.] Estimate the differential entropy as the logarithm of the estimated volume:
\begin{equation}\label{Eq:hMa}
{\hat h}_{Ma} =  d \log_2(r) +\log_2(Q(r)).  
\end{equation}
\end{itemize}

Note that, according to the definition of $C(r)$, in Eq. \ref{Eq:C}, it can be related to $Q(r)$ as $C(r)=\frac{1}{Q(r)}$, and Eq. \ref{Eq:hMa} can be rewritten as
\begin{equation}\label{Eq:hMa2}
{\hat h}_{Ma} =  d \log_2(r) -\log_2(C(r)).  
\end{equation}

Comparing Eq. \ref{Eq:hMa2} to Eq. \ref{EqID2} we conclude that the $h$ in Eq. \ref{EqID2} is the Ma's entropy estimate, ${\hat h}_{Ma}$. As a consequence, the line fitting procedure explained for the ID estimation can also be used for DE estimation, where slope and $y$-intercept parameters play the role of ID and DE estimates, respectively. On the other hand, the ambiguity problem mentioned in Section \ref{Sec:IDestim} is crafted into a tool that allows for multiscale analysis through a perspective similar to that proposed in \cite{Little2009}, where almost linear segments with different angular and linear coefficients give clues regarding the structure of the underlying manifold.

As an illustration of this multiscale analysis, we consider again results shown in Figure \ref{Fig03}, with two almost linear intervals. The estimated line segments have angular coefficients close to 1 and 2, respectively, associated to DE estimates $h_1 \approx -0.65$, thus close to the theoretical DE of the source, $-0.67$ bits, and $h_2 \approx 2.7$ bits, which is close to the logarithm of the {\em ring} length in Fig. \ref{Fig01}, $\log_2 (2 \pi) \approx 2.65$ bits.

In other words, the two almost linear segments suggest that (a) at small scales the dominant structure is 2D, with an effective area close to $2^{h_2}$, whereas (b) at larger scales the dominant structure becomes roughly 1D, with effective length close to $2^{h_1}$.

\section{A method for visual analysis of ID and DE}
\label{Sec:Proposal}

The method proposed here is a straightforward recombination of the approaches explained in Sections \ref{Sec:IDestim} and \ref{Sec:DEestim}, chosen for their simplicity and data efficiency (for both methods consider all possible pairs of observations). In this recombination, it is assumed that:
\begin{itemize}
\item ID is constant over the variable domain.
\item Probability density function is locally uniform.
\end{itemize}
\noindent The method is organized in 7 steps. The first 5 steps are presented below, whereas the remaining ones are presented in Subsection \ref{Sec:Scarcity}, where the bias problem is addressed.

\begin{itemize}
\item[(S1)] Compute the supremum norm for each vector ${\bf x}(i)-{\bf x}(j)$, $i<j$. Double each norm and store the results in an array ${\bf r}$.

\item[(S2)] Sort ${\bf r}$. Now ${\bf r}(k)$ is the edge size of the hypercube that yields $k$ coincidences.

\item[(S3)] Plot $\log_2(k/L_r)$ versus $\log_2({\bf r}(k))$, where $L_r$ is the length of the array ${\bf r}$ and $k$ ranges from 1 to $L_r$ (optionally, points can be resampled and interpolated for better visualization).

\item[(S4)] Plot ID hypotheses $\log_2({\bf r}(k))$ versus $d \log_2({\bf r}(k))$ for some arbitrary $d < D$.

\item[(S5)] Visually chose IDs, $\hat d$, and DEs, $\hat h$, of selected segments of the plot (segments where the slope can be approximated by a constant).

\end{itemize}

{\bf Example:} Let ${\bf X} = \{ {\bf x}(1), ~{\bf x}(2), ~{\bf x}(3), ~{\bf x}(4), ~{\bf x}(5) \}$ be a set of $N=5$ independent observations of a random source, namely:

${\bf x}(1)=[92,~46,~138]$,

${\bf x}(2)=[4,~ 2,~ 7]$,

${\bf x}(3)=[48,~24,~72]$,

${\bf x}(4)=[26,~13,~40]$,

${\bf x}(5)=[41,~21,~62]$.

\noindent Supremum norms for all 10 non-redundant observation pairs are computed and multiplied by 2, yielding ${\bf r} =$ [262, 132, 196, 152, 130, 66, 110, 64, 20, 44].  

These values are sorted in ascending order as: 
\[
{\bf r} = [20,~ 44,~   64,~    66,~    110,~    130,~    132,~    152,~    196,~    262].
\] 

\noindent Thus, ${\bf r}(4)=66$, for instance, means that a cube of edge 66 around each observation yields 4 coincidences. For this particular value we can compute $C(66)$ as the number of coincidences (4) divided by the total number of pairs (10), yielding the pair $(\log_2(66),~\log_2(4/10)) \approx (6.0,~-1.3)$ to be plotted.

\begin{figure}[htb]
\centering{\includegraphics[width=70mm]{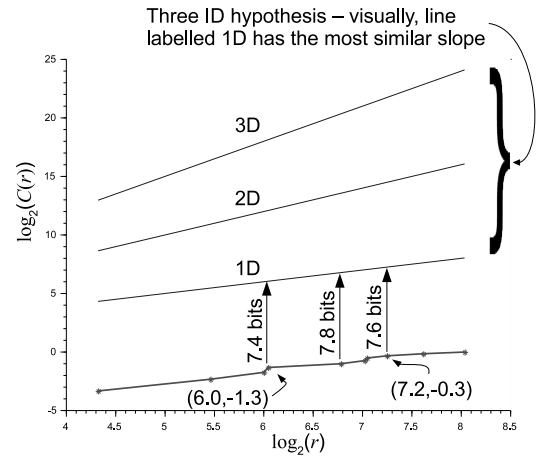}}
\caption{Plot of ordered pairs $\left(\log_2(r),~\log_2 C(r)\right)$ . The resulting plot is visually compared to 3 ID hypothesis. The best match is 1D (thus ${\hat d}=1$), and the average vertical distance from plotted points to the corresponding line yields an estimated ${\hat h} \approx 7.6$ bits.}
\label{Fig04}
\end{figure}

Proceeding likewise for all values in vector $\bf r$, the plot in Fig. \ref{Fig04} is obtained. Through visual inspection, it is possible to infer that observations roughly lie in an 1D structure, for the candidate with most similar slope in Fig. \ref{Fig04} equals one. In other words, although observation are given in $D=3$, we are able to infer that they lie in manifold whose intrinsic dimension is $d=1$. 

Besides, once $d$ is estimated, the DE can be estimated as the average value of differences $d\log_2({\bf r}(k))-\log_2 C({\bf r}(k))$. In this example, the differences for three arbitrarily chosen points are $7.4,~7.8$ and $7.6$, thus yielding an average DE estimate of $\hat h = 7.6$ bits.

These estimates for ID and DE suggest that the five observations in this example were sampled from an 1D structure of length $2^{\hat h} \approx 194$. Indeed, the $N=5$ points were uniformly drawn from a noisy linear segment with length $\sqrt{100^2+50^2+150^2} \approx 187$. Therefore, the ID of the underlying 1D manifold was correctly inferred, while its length was roughly guessed through the estimated DE.

\subsection{Bias compensation}
\label{Sec:Scarcity}

Both ID and DE estimators combined in this work are based on the exponentially growing fraction of patterns randomly coinciding, on average, inside small hypercubes of growing edge. Ideally, this edge should be vanishingly small, but in practice the number of observations is finite, what yields two antagonistic restrictions, namely: that the hypercube size should be as small as possible, thus containing a small fraction of observations, and that this fraction should be as large as possible, for statistical reasons. 

In \cite{Smith1988} it is shown that ID is always underestimated by Eq. \ref{EqID} in the simple case of a hypercube inside which the probability density of a pattern being observed is uniform, even for an unlimited amount of data.
The equations in \cite{Smith1988} that explain this bias are rewritten here as Eq. \ref{Eq:Smith1} and \ref{Eq:Smith2} for the reader convenience:
\begin{equation}\label{Eq:Smith1}
C_0(r)=(r(2-r))^d
\end{equation}
\begin{equation}\label{Eq:Smith2}
d_0(r)=d \times \left( 1 - \frac{r}{2-r} \right)
\end{equation}
\noindent where $C_0(r)$ and $d_0(r)$ stands for theoretical estimates of $C$ and $d$ for an RV uniformly distributed in a $d$-dimensional hyper-cube of edge $r$. It is noteworthy that $d_0(r)$ is the derivative of $\log C_0(r)$ with respect to $\log r$.

In \cite{Smith1988}, under the following arbitrary restrictions: 
\begin{itemize}
\item{R1:} $d_0(r) \geq 0.95 d$, which imposes an estimate deviation tolerance, and 
\item{R2:} the minimum $r$ is $1/4$ of the maximum $r$, which allows the expected exponential proportionality of Eq. \ref{Eq:Smith1} to appear,
\end{itemize}
\noindent it is shown that the minimum number of observations, $N_{min}$, for a proper ID estimation depends upon the true ID, $d$, as
\begin{equation}\label{Eq:Smith}
N_{min} \approx 42^d.
\end{equation}

This requirement is impractical for most real applications. For instance, even for $d$ as low as 5 an experimenter would need more than 100 million independent samples in order to obtain a good ID estimate. In subsequent works this result was replaced with less restrictive ones such as in \cite{Eckmann1992}, where a much simpler analytic model is used, yielding 
\begin{equation}\label{Eq:Eckmann}
N_{min} \approx 10^{d/2}
\end{equation}

\begin{figure}[htb]
\centering{\includegraphics[width=70mm]{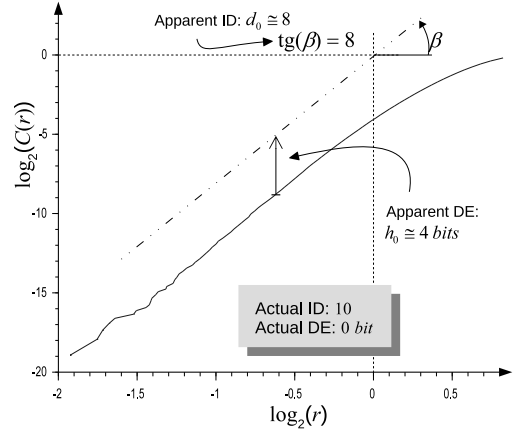}}
\caption{An instance of biased estimates for ID and DE. The actual ID and DE are 10D and 0 bit, but the visual analysis yields estimates around 8D and 4 bits, respectively. These strong biases are caused by the smallness of the dataset, as compared to its actual ID.}
\label{FigFalsaID}
\end{figure}

In spite of their differences, both works agree that small datasets yield false ID estimates, biased toward lower values. For instance, with $N=1000$ observations independently and uniformly sampled in a 10D hypercube, the visual approach used here yields the result presented in Fig. \ref{FigFalsaID}, suggesting a wrong ID estimate of about 8D, as well as a wrong DE estimate of about 4 bits (the actual DE is 0 bit). 

To predict and compensate for both biases, we developed an approach built upon the analytical model proposed in \cite{Smith1988}. In practical terms, it consists of completing a table of underestimated IDs, for a given $N$, then using this table to infer the unbiased ID, which in turn allows the estimation of a bias compensation for the DE too.   

The above mentioned table is based on Eqs. \ref{Eq:Smith1} and \ref{Eq:Smith2} and on a coarse estimation of the average supremum distance from an observation to its nearest neighbour, $\bar r$, where for $N$ observations over a regular grid in a $d$-dimensional unit volume hypercube, one should expect 
\begin{equation} \label{Eq:remedio}
{\bar r}(N,d)=\frac{1}{1+N^{1/d}}.
\end{equation}

To obtain this average supremum distance we first consider a line segment of unit length which is equally split into $n+1$ intervals, thus allowing the placement of $n$ equally spaced points apart from each other by $r=1/(1+n)$. Likewise, in a unit area square, $N=n^2$ points can be regularly arranged by keeping the same $r$ (as a result of the same $n=N^{1/2}$) as the supremum distance between neighboring points. Through the generalization of this simple reasoning for a unit volume hypercube of dimension $d$, where $N=n^d$ points can be regularly arranged in the vertices of a grid, $r$ remains the supremum distance between any neighboring points of this grid. Therefore, given $N$ and $d$, there is at least one arrangement of the $N$ points separated from nearest neighbors by $r=1/(1+N^{1/d})$. On the other hand, for $N$ points randomly placed inside that same $d$-dimensional hypercube, the supremum distance between neighboring points is a random variable, say $R$, but if its underlying probability density function is uniform, we can use Eq. \ref{Eq:remedio} as a coarse approximation of the expected value for $R$.


This approximation experimentally proved to be useful for $N << 2^d$, which tends to be the case for high ID values, were bias correction is even more relevant. For instance, if $d=10$ and $N=100$, the prediction is ${\bar r}(100,10) \approx 0.38$, which is the same value experimentally obtained up to two decimal places. Likewise, if $d=20$ and $N=10000$, the prediction is ${\bar r}(10000,20) \approx 0.37$, whereas the experimental value is about $0.39$. By contrast, for less sparse datasets, such as for $d=5$ and $N=100$, the prediction is ${\bar r}(100,5) \approx 0.21$, whereas the experimental value is about $0.28$.




Applying Eq. \ref{Eq:remedio} to Eq. \ref{Eq:Smith2} we obtain 
\begin{equation}\label{Eq:IDbias}
d_0(N,d)=d \times \left( 1 - \frac{{\bar r}(N,d)}{2-{\bar r}(N,d)} \right)
\end{equation}

By definition \cite{Cover2012}, a random variable with uniform probability density inside a hypercube of unitary volume has null differential entropy. Therefore, given that Smith's bias is calculated precisely for this random variable, Eq. \ref{Eq:hMa} should yield $h_{Ma}=0$ in this case, and any imbalance between $\log_2(C_0(r))$ and $d_0(r) \log_2(r)$ is to be taken as an entropy bias, $\Delta h$. Therefore, for the estimated $d_0$ the expected DE bias is
\begin{equation}\label{Eq:hhat}
\Delta h=\log_2{C_0(r)}-d_0(r) \log_2(r)
\end{equation}
\noindent Applying Eq. \ref{Eq:Smith1} and \ref{Eq:Smith2} to Eq. \ref{Eq:hhat} we obtain
\[
\Delta h=\log_2{(r(2-r))^d}-d \left( 1 - \frac{r}{2-r} \right) \log_2(r)
\]
\noindent which can be simplified to
\begin{equation}\label{Eq:passagem}
\Delta h=d \left( \left(\frac{r}{2-r}\right) \log_2(r)+\log_2{(2-r)} \right)
\end{equation}
\noindent Using  Eq. \ref{Eq:remedio} into Eq. \ref{Eq:passagem}, we obtain
\[
\Delta h(N,d)  =
\]
\begin{equation}\label{Eq:DEbias}
 d \left( \left(\frac{{\bar r}(N,d)}{2-{\bar r}(N,d)}\right) \log_2({\bar r}(N,d))+\log_2{(2-{\bar r}(N,d))} \right)
\end{equation}

Finally, to compensate for biases, Steps S1 to S5, as proposed in Section \ref{Sec:Proposal}, are followed by two more steps, namely:
\begin{itemize}
\item[(S6)] Using Eq. \ref{Eq:IDbias}, find the compensated ID estimate, ${\bar d}$, that yields the closest $d_0(N,d)$ to the visually estimated $\hat d$.
\item[(S7)] Obtain $\Delta h(N,{\bar d})$ using Eq. \ref{Eq:DEbias} and compute a compensated DE estimate as ${\bar h}=\hat h-\Delta h(N,{\bar d})$.    
\end{itemize}

{\bf Illustration:} An experimenter gathered $N=1000$ multivariate observations with $D=20$ attributes, and this observer applies the visual method (steps S1 to S5), thus obtaining the solid curve in Fig. \ref{FigFalsaID}. A naive experimenter would believe that the ID of that data is 8, according to the angle of the dashed line (found after visual comparison between some  competing slopes). Lets call it the {\em apparent ID}, $\hat d \approx 8$, associated to the {\em apparent DE}, $\hat h \approx 4$ bits. However, because $N$ is too small as compared to $42^8$ \cite{Smith1988}, or even to $10^{8/2}$ \cite{Eckmann1992}, one should not accept the result of this first analysis. Proceeding with step S6, a range of possible IDs near $\hat d$ is considered and Eq.\ref{Eq:IDbias} is used to complete Table \ref{Tab:bias}, from which it is possible to infer that the {\em apparent ID} near 8 corresponds to a bias compensated ID of $10$, which is the actual ID of the data source used in this illustration. On step S7, Eq. \ref{Eq:DEbias} further yields $\Delta h(1000,10) \approx 4.2$ and a less biased DE estimate is finally obtained as ${\hat h}-\Delta h(1000,10) \approx -0.2$ bits (the actual DE of the data source used in this illustration is zero).

\begin{table}[htb]
\centering
\caption{Bias compensation table for Figure \ref{FigFalsaID}.}\label{Tab:bias}
\begin{tabular}{|c|c|c|c|c|c|c|c|}
\hline 
$d$ & 8  & 9  & \bf 10 & 11 & 12\\ 
\hline 
$d_0(1000, d)$ & 6.6 & 7.3 & \bf 7.99 & 8.69 & 9.36\\
\hline 
\end{tabular} 
\end{table}

\section{Experimental results}
\label{Sec:Results}

Two sets of experimental results are presented. First with two artificial data whose intrinsic dimensions are known, and their corresponding results are presented as evidences in favor of the proposed approach. Those results palliate the difficulty of providing statistical analysis for the method, since it depends upon visual (human) evaluation as part of the process. Two real datasets are analyzed afterwards, and despite the fact that their intrinsic dimensions  were already analyzed in former published papers, our results induce some interesting questions regarding estimates consistency and the need for bias compensation.

\begin{figure}[htb]
\centering{\includegraphics[width=70mm]{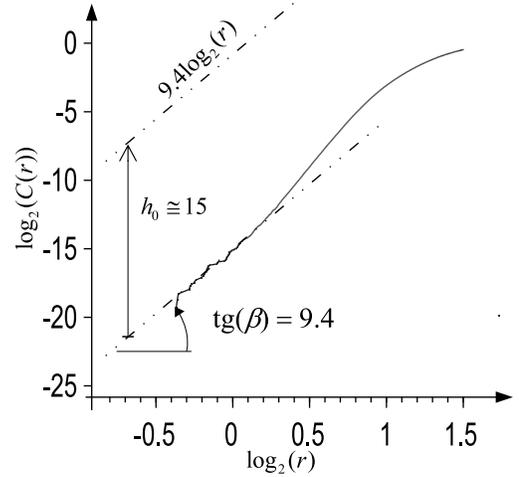}}
\caption{ID estimation for a 12-dimensional manifold in 72-dimensions proposed in \cite{Hein2005}. For values of $\log_2(r)$ from -0.3 to 0.1 (small observation scale), the apparent ID is about 9.4, whereas the apparent DE is about 15 bits, both biased.}
\label{FigHein1}
\end{figure}

\begin{figure}[htb]
\centering{\includegraphics[width=70mm]{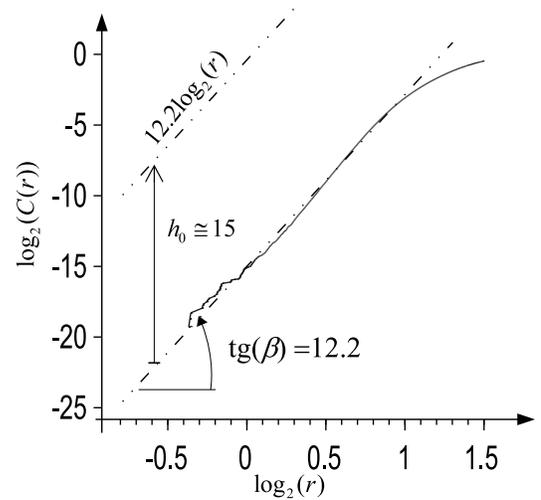}}
\caption{ID estimation for a 12-dimensional manifold in 72-dimensions proposed in \cite{Hein2005}. For values of $\log_2(r)$ from 0.1 to 1.0 (coarse observation scale), the apparent ID is about 12.2, whereas the apparent DE remains around 15 bits.}
\label{FigHein2}
\end{figure}

The first artificial dataset source corresponds to a 12-dimensional manifold ($d=12$) in 72-dimensions ($D=72$) first proposed in \cite{Hein2005}, then reused afterwards in \cite{Rozza2012} and \cite{Einbeck2019}, which makes it a suitable dataset for comparison purposes. $N=1600$ random data points were used and two results are separately presented in Figures \ref{FigHein1} and \ref{FigHein2} for a better visualization of an interesting aspects of this dataset.
For values of $\log_2(r)$ from -0.3 to 0.1 (fine observation scale), the apparent ID is about 9.4, whereas the apparent DE is about 15 bits, as can be better observed in Fig. \ref{FigHein1}. As for Fig. \ref{FigHein2}, we observe instead an  apparent ID of about 12.2, whereas the apparent DE remains around 15 bits, for values of $\log_2(r)$ from 0.1 to 1.0 (coarse observation scale)\footnote{The visual comparison to slopes such as 9.4 and 12.2 was induced by the values found in Table \ref{Tab:biasHein}, for integer values of compensated IDs.}.

Again, proceeding with step S6, a range of possible IDs is considered and Eq.\ref{Eq:IDbias} is used to complete Table \ref{Tab:biasHein}, from which one may conclude that for fine scales of observation, the actual ID of the corresponding manifold is about 12 (from 9.4, after bias compensation), whereas its biased DE of about 15 bits should be compensated (step S7) to  ${\hat h}-\Delta h(1600,12) \approx 15-4.8 = 10.2$ bits. It is noteworthy that 12 is indeed the artificially imposed ID to the manifold underlying this dataset. Moreover, in \cite{Hein2005} it is highlighted that this manifold has a  {\em ``high curvature and nontrivial probability measure effects on the manifold''}, and we believe that the second linear trend shown in Fig. \ref{FigHein2} is a consequence of that high curvature, for the apparent ID of about 12.2 is compensated to 16, which is compatible with the idea that a 12D manifold can be curved to the point that, for a coarse observation scale, it forms a (hollow) structure of dimension higher than 12. The biased DE of such structure is compensated to  ${\hat h}-\Delta h(1600,16) \approx 15-5.8 = 9.2$ bits.

\begin{table}[htb]
\centering
\caption{Bias compensation table for figures \ref{FigHein1} and \ref{FigHein2}.}\label{Tab:biasHein}
\begin{tabular}{|c|c|c|c|c|c|c|c|}
\hline 
$d$ & 11  & \bf 12  &  13 & 14 & 15 & \bf 16\\ 
\hline 
$d_0(1600, d)$ & 8.8 & \bf 9.4 &  10.1 & 10.8 & 11.5 & \bf 12.2\\
\hline 
\end{tabular} 
\end{table}

\begin{figure}[htb]
\centering{\includegraphics[width=70mm]{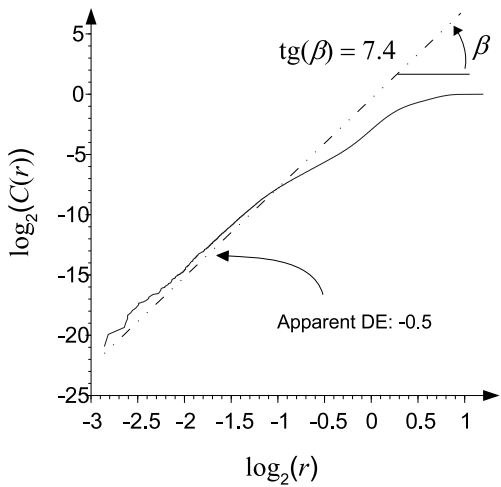}}
\caption{ID estimation for the dataset labeled ``Data Set D''. For small values of $log_2(r)$ the apparent ID is about 7.4, whereas the apparent DE is close to zero, about -0.5 bits.}
\label{FigSantaFeD}
\end{figure}

The second artificial dataset is labeled ``Data Set D'' \cite{Gershenfeld1993}, also used in \cite{Rozza2012} under label ``Santa Fe dataset''. As explained in \cite{Gershenfeld1993}, it corresponds to a {\em ``relatively long series of known high-dimensional dynamics (...) with weak nonstationarity''} with 100,000 points obtained by numerical integration of the equations of motion for a damped, driven particle. 
We organized the simulated values in $N=2000$ 50D patterns, as in \cite{Rozza2012}, which yielded the visual result presented in Fig. \ref{FigSantaFeD}, where an apparent ID of about 7.4 is observed, along with a small apparent bias of about -0.5 bits\footnote{The visual comparison to slopes such as 6.7, 7.4 and 8.1 was induced by the values found in Table \ref{Tab:biasSantaFeD}, for integer values of compensated IDs.}.   

Consulting Table \ref{Tab:biasSantaFeD}, one may infer a compensated ID of about 9 (the actual ID of this artificial data source), and a corresponding compensated DE of about ${\hat h}-\Delta h(2000,9) \approx -0.5-4.1 = -4.6$ bits.  

\begin{table}[htb]
\centering
\caption{Bias compensation table for figure \ref{FigSantaFeD}.}\label{Tab:biasSantaFeD}
\begin{tabular}{|c|c|c|c|c|c|c|c|}
\hline 
$d$ & 7  &  8  & \bf 9 & 10 & 11\\ 
\hline 
$d_0(2000, d)$ & 6.0 &  6.7 &  \bf 7.4 & 8.1 & 8.8\\
\hline 
\end{tabular} 
\end{table}

The first real dataset used in this paper is labeled ``Paris-14E Parc Montsouris'' in \cite{Camastra2009}, corresponding to a time series formed by daily average temperatures (in tenths of Celsius degrees) in Paris, from January 1, 1958 to December 31, 2001. We organized the 15,706 measurements in $N=785$ patterns of $D=20$ measurements each. In \cite{Camastra2009} three ID estimation algorithms were applied to this dataset, including GP, with which the authors of \cite{Camastra2009} estimated an ID of 4.91.  

\begin{figure}[htb]
\centering{\includegraphics[width=70mm]{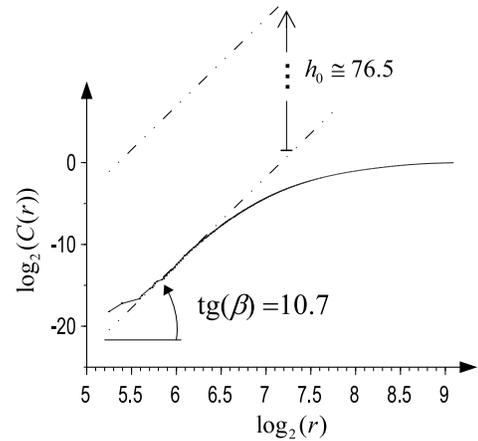}}
\caption{ID estimation for the dataset labeled ``Paris14e Parc Montsouris''. For values of $\log_2(r)$ from 5.7 to 6.3 the apparent ID is about 10.7, whereas the apparent DE is about 76.5 bits.}
\label{FigMontsouris}
\end{figure}

By contrast, Figure \ref{FigMontsouris} presents our reproduction of the experiment with the Grassberger-Procaccia approach, where for values of $\log_2(r)$ from 5.7 to 6.3 the apparent ID is about 10.7, whereas the apparent DE is about 76.5 bits.

This visual result, even before any bias compensation, suggests that an ID of about 5 is far from any ID value estimated for small values of $\log_2(r)$. We then conjecture that the authors of \cite{Camastra2009} estimated an average slope for a wide range of $\log_2(r)$, which indeed would yield an ID estimate near 5. Besides, in \cite{Rozza2012} twelve different ID estimators were applied to this same dataset, yielding inconsistent estimates ranging from 3.71 up to 13.52. 

In this work, we assume that the apparent ID of 10.7 in Fig. \ref{FigMontsouris} as our best guess for small values of $r$, whose bias compensation, according to Table \ref{Tab:biasMontsouris} yields an ID of about 14. Likewise, the corresponding compensated DE is about ${\hat h}-\Delta h(785,14) \approx 76.5-5.1 = 71.4$ bits. 

\begin{figure}[htb]
\centering{\includegraphics[width=70mm]{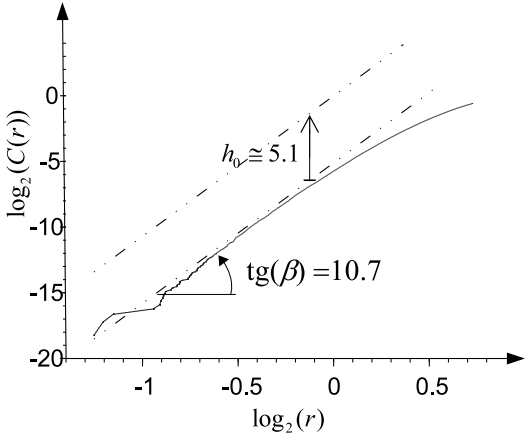}}
\caption{ID estimation for an artificial 14D dataset of null DE. The apparent ID of about 10.7 is visually compatible with Fig. \ref{FigMontsouris}, whereas the apparent DE of about 5.1 bits equals the DE bias compensation applied to the ``Paris14e Parc Montsouris'' dataset.}
\label{FigMontsouris_test}
\end{figure}

To check this result, we did an additional analysis similar to that shown in Fig. \ref{FigFalsaID}, this time with $N=785$ random observations of a random variable uniformly distributed in a hyper-cube of 14 dimensions, thus with actual ID of 14, and actual DE of 0 bit. In this experiment, the apparent ID and DE were found to be $d_0=10.7$ and $h_0 \approx 5.1$ bits, as shown in Fig. \ref{FigMontsouris_test}, which seems to confirm that results shown in Fig. \ref{FigMontsouris} are compatible with a random source of 14D (apparent ID of about 10.7), to which a bias compensation of about 5.1 bits is necessary. In other words, Fig. \ref{FigMontsouris_test} corroborates the idea that the ``Paris14e Parc Montsouris'' dataset lies in a 14D (thus greater than 10.7) manifold whose DE is about 71.4 (instead of 76.5) bits.

\begin{table}[htb]
\centering
\caption{Bias compensation table for figure \ref{FigMontsouris}.}\label{Tab:biasMontsouris}
\begin{tabular}{|c|c|c|c|c|c|c|c|c|}
\hline 
$d$ & 12  &  13  & \bf 14 & 15 & 16 & 17\\ 
\hline 
$d_0(785, d)$ & 9.3 &  10.0 &  \bf 10.7 & 11.4 & 12.0 & 12.7\\
\hline 
\end{tabular} 
\end{table}

\begin{figure}[htb]
\centering{\includegraphics[width=70mm]{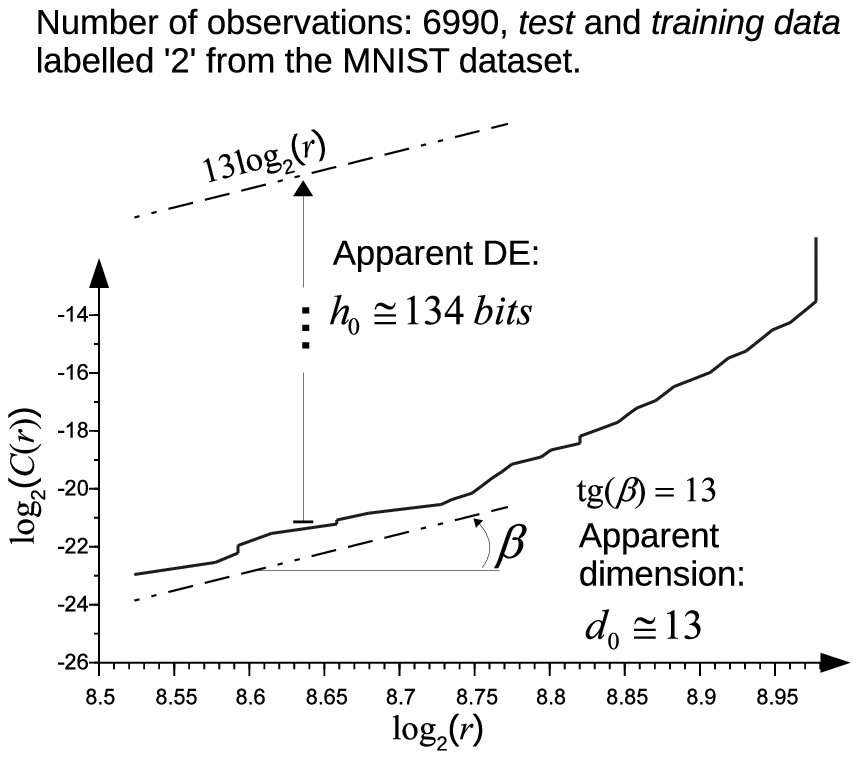}}
\caption{ID estimation for digit 2 from the MNIST dataset. The apparent ID is about 13, whereas the apparent DE is about 134 bits, both biased.}
\label{FigFalsaID13}
\end{figure}

Another experiment with real data was done with all $N=6990$ available observations of digits labeled `2' in the MNIST dataset\cite{Lecun1998}, for practical purposes, we label this dataset as ``MNIST 2''. Digit `2' was chosen to allow a comparison of our result to similar experiments reported in \cite{Costa2004Learning}, \cite{Facco2017} and \cite{Hein2005}. Figure \ref{FigFalsaID13} corresponds to the visual analysis from this experiment, where an apparent ID of 13 was observed, associated to an apparent DE of about 134 bits.

The visually estimated ID around 13 is in agreement to results presented in \cite{Costa2004Learning}, \cite{Facco2017} and \cite{Hein2005}, but it seems to be a misleading observation, for the corresponding bias-compensated ID is higher than 13. Indeed, after going through steps S6 and S7, Table \ref{Tab:bias13} suggests that, for $N=6990$ observations, an apparent ID of about 13 is expected when the actual ID is 17.

\begin{table}[htb]
\centering
\caption{Bias compensation table for Figure \ref{FigFalsaID13}.}\label{Tab:bias13}
\begin{tabular}{|c|c|c|c|c|c|c|c|}
\hline 
$d$ & 13 & 14 & 15  & 16  &  \bf 17 & 18\\ 
\hline 
$d_0(6990,d)$ & 10.3 &  11.0 & 11.7 & 12.4 & \bf 13.1 & 13.7\\
\hline 
\end{tabular} 
\end{table}

Besides, by assuming that the actual ID is 17,  Equation \ref{Eq:DEbias} predicts a DE bias of about $\Delta h(6990,17)=6.4$ bits, therefore, we estimate that the actual DE is $h_0-\Delta h(6990,17)=134-6.4 \approx 128$ bits. This is less than the DE estimated by \cite{Costa2004Learning}, of about 145 bits. Such a discrepancy may be partially accounted for the fact that in \cite{Costa2004Learning}, the estimated DE is the intrinsic R\'enyi $\alpha$-entropy for $\alpha=1/2$, whereas we estimate the collision DE ($\alpha=2$).

As in the former experiment with real datasets, to check our results, we did an additional analysis similar to that shown in Fig. \ref{FigFalsaID} with $N=6990$ random observations of a random variable uniformly distributed in a unit-volume hyper-cube of 17 dimensions, thus with actual ID and DE equal to 17 and 0 bits, respectively. In this experiment, the apparent dimension and entropy were found to be $d_0=13$ and $h_0 \approx 6.8$ bits, with a visual aspect quite similar  to those presented in Figures \ref{FigFalsaID} and \ref{FigMontsouris_test}. This result seems to confirm our conclusion that ``MNIST 2'' samples lies in a 17D manifold. However, the bias compensation prediction of about 6.4 slightly deviated from the observed bias of about 6.8 bits, for the artificial data used in the test. 

\section{Conclusion}
\label{Sec:Conclusions}

A new approach for bias-compensated estimation of intrinsic dimension and differential entropy was proposed in this paper. It corresponds to the natural combination of previously published estimation methods, one for {\em collision entropy} -- or quadratic entropy --, by Ma \cite{Ma1981}, and another for {\em correlation dimension}, by Grassberger and Procaccia \cite{Grassberger1983_1}. In the first part of this work it was explained why these two approaches are connected in spite of their different goals, and how ID and DE should be regarded as two complementary aspects of random observations analysis, thus yielding a joint estimation approach.

An important aspect of this approach is its dependency on scale of analysis. Although it is frequently regarded as a practical obstacle for estimators, we propose that estimates at different scales convey different perspectives of underlying manifolds. Accordingly, we propose a pragmatic visual approach, followed by some illustrations.     

On the other hand, the seminal work by Smith \cite{Smith1988} is a clear warning regarding the always present bias in the Grassberger-Procaccia estimator. Then, we built upon the theoretical model used by Smith to introduce a systematic bias compensation for both ID and DE estimation, whose use is validated through experiments with real data and further illustrated through experiments with synthetic ones. 

It is worthy noticing that while this work is strongly based on Smith's analysis, which yields a quite severe restriction on the minimum number of observations for a reliable estimate of $d$, as pointed out in Eq. \ref{Eq:Smith}, that restriction does not apply to this work. Indeed, Smith's analysis imposes that the estimated dimension should not be less than 95\% of the actual one, without any kind of bias compensation. By contrast, in this work, instead of imposing a bias threshold, we use Smith's formula to compensate for that bias, even if the number of observations is much less than $N_{min} \approx 42^d$.

The proposed approach is developed under the assumptions that the ID is constant over the variable domain and that the underlying probability density function is locally uniform. If these assumptions are not verified, the proposed approach should not be applied. Notwithstanding, thanks to the visual analysis that is an important part of this approach, and taking into account its potential for a geometrical analysis of manifolds as a whole, as proposed in \cite{Montalvao2019}, we believe that the study of visual patterns (of log(r) versus log(C)) even when these assumptions are violated can be a promising research subject for the future.

We also believe that the proposed tool for manifold analysis can be useful in pattern recognition context, specially in this renewed era of artificial neural network applications. Indeed, many researchers concerned with this topic seem to converge to the conclusion that relevant insights should come from the study of manifolds. In this work, we try to provide a pragmatic tool for the bias-corrected estimation of manifold volume and intrinsic dimension. This can be regarded as a first step in understanding how layered processing structures disentangle data manifolds, and how to eventually improve it.


%

\ifCLASSOPTIONcaptionsoff
  \newpage
\fi

\section*{Acknowledgments}
This work has been supported by The Conselho Nacional de Desenvolvimento Cient\'ifico e Tecnol\'ogico (CNPq) to J.M., grant 304853/2015-1 and 308319/2018-4.

\bibliographystyle{IEEEtran}

\bibliography{IEEEabrv,so_biblio}

\end{document}